\documentclass[letterpaper, 10 pt, conference]{ieeeconf}  

\IEEEoverridecommandlockouts                              

\usepackage{graphicx} 
\usepackage{amsmath} 
\usepackage{amssymb}  

\usepackage{booktabs,tabularx}
\usepackage{subcaption}

\title{\LARGE \bf
Knowledge-Guided Manipulation Using Multi-Task Reinforcement Learning
}

\author{
Aditya Narendra$^{1}$,
Mukhammadrizo Maribjonov$^{2}$,
Dmitry Makarov$^{3}$,
Dmitry Yudin$^{4}$,
and Aleksandr Panov$^{5}$%
\thanks{$^{1}$Aditya Narendra is with MIRAI, Moscow, Russia and MBZUAI, Abu Dhabi, UAE
        {\tt\small  aditya.narendra@mbzuai.ac.ae}}%
\thanks{$^{2}$Mukhammadrizo Maribjonov is with MIRAI, Moscow, Russia and Innopolis University, Russia
        {\tt\small m.maribjonov@innopolis.university}}
\thanks{$^{3}$Dmitry Makarov is with MIRAI, Moscow, Russia
        {\tt\small makarov.d@miriai.org}}%
\thanks{$^{4}$ Dmitry Yudin is with MIRAI and AXXX, Moscow, Russia
        {\tt\small yudin.da@miriai.org, yudin@cogailab.com}}
\thanks{$^{5}$ Aleksandr Panov is with MIRAI and AXXX, Moscow, Russia
        {\tt\small panov.a@miriai.org, panov@cogailab.com}}
}

\begin{document}

\bstctlcite{IEEEexample:BSTcontrol}

\maketitle
\thispagestyle{empty}
\pagestyle{empty}


\begin{abstract}
This paper introduces Knowledge-Guided Massively Multi-task Model-based Policy Optimization (KG-M3PO), a framework for multi-task robotic manipulation in partially observable settings that unifies Perception, Knowledge, and Policy. KG-M3PO leverages a model-based policy optimization method to control backbone with an online 3D scene graph that grounds open-vocabulary detections into a metric, relational representation. A dynamic-relation mechanism updates spatial, containment, and affordance edges at every step, and a graph neural encoder is trained end-to-end through the RL objective so that relational features are shaped directly by control performance. Multiple observation modalities (visual, proprioceptive, linguistic, and graph-based) are encoded into a shared latent space, upon which the RL agent operates to drive the control loop. The policy conditions on lightweight graph queries alongside visual and proprioceptive inputs, yielding a compact, semantically informed state for decision making.

Experiments on a suite of manipulation tasks with occlusions, distractors, and layout shifts demonstrate consistent gains over strong baselines: the knowledge-conditioned agent achieves higher success rates, improved sample efficiency, and stronger generalization to novel objects and unseen scene configurations. These results support the premise that structured, continuously maintained world knowledge is a powerful inductive bias for scalable, generalizable manipulation: when the knowledge module participates in the RL computation graph, relational representations align with control, enabling robust long-horizon behavior under partial observability.

\end{abstract}

\section{Introduction}

Robotic manipulation in unstructured, everyday environments is a fundamentally partially-observable problem. Critical information is routinely missing: objects are occluded, moved by prior actions, or concealed inside containers; lighting and viewpoints changecog continuously. In these conditions, agents must maintain and reason over a persistent world model. Purely reactive policies that map pixels to torques are prone to state aliasing and fail at long-horizon credit assignment, while traditional model-based approaches often rely on brittle, hand-tuned geometric pipelines. The practical need for \emph{multi-task} competence (e.g., pick, place, open) and the desire for \emph{generalizable methods} that can be applied across different robotic embodiments (e.g., Franka and UR5 robots) further compounds the challenge, necessitating a unified architectural framework for learning.\\

\begin{figure}[t]
    \centering
    \includegraphics[width=0.6\columnwidth]{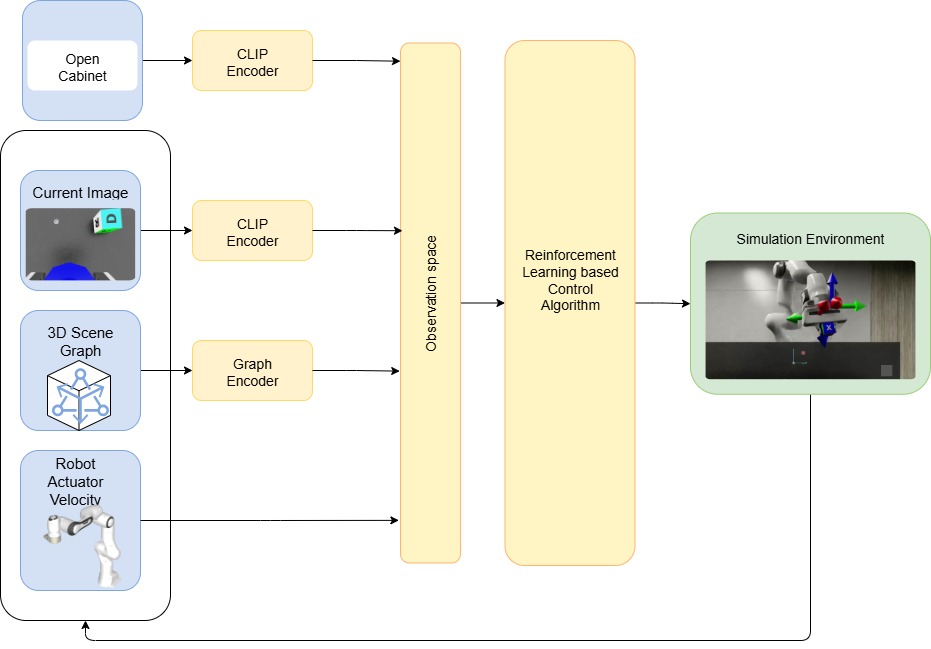}
    \caption{Overview of the end-to-end training pipeline for KG-M3PO (M3PO augmented with an online KG encoder trained end-to-end).   
    Multiple observation modalities (language goal, current image, 3D scene graph, and proprioception) are encoded into a common observation space.  
    A reinforcement learning algorithm (in our case, M3PO) then drives the control loop inside the simulation environment.  
    The knowledge encoder (graph) is trained directly through the RL loss.}
    \label{fig:train-pipeline}
\end{figure}

It is clear that a scalable solution requires the integration of three key components: (i) \emph{structured world knowledge} that persists across time steps, (ii) a \emph{policy learning} framework that leverages this knowledge to disambiguate partial observations, and (iii) \emph{multi-task generalization} to learn generalizable representations for a given robot. To this end, we present a suite of manipulation tasks on two distinct manipulators (Franka and UR5), spanning both fully and partially observable settings. For each robot, we train both single-task expert agents as well as multi-task generalized agents. The fully observable regime tests whether structured knowledge can improve sample efficiency even when the state is technically visible. The partially observable regime introduces tasks that are inherently challenging for camera-only policies, such as retrieving an object hidden behind an occluding wall or inside a container, unless the agent maintains a dynamic, structured state estimate. Our goal is to develop a \emph{knowledge-guided framework} for learning multi-task RL agents that bridges this perception-to-policy gap. \\

Translating perception into reliable manipulation policies under partial observability presents several interconnected challenges:
\begin{itemize}
    \item \textbf{State Estimation:} Single-view visual features are insufficient under fluctuating visibility; the agent must integrate evidence over time to maintain beliefs over object poses, containment states, and affordances.
    \item \textbf{Relational Reasoning:} Effective control relies on understanding spatial and functional \emph{relations} (e.g., `on', `in', `near', `behind', `under'). These relations are inherently graph-structured, change dynamically with actions, and must be kept temporally consistent.
    \item \textbf{Multi-Task Representation Learning:} A single agent for a given robot must acquire representations that support heterogeneous goals and object sets without catastrophic forgetting, while remaining sample-efficient. 
    \item \textbf{Credit Assignment:} Learning must bridge perception and control; improvements in state estimation must translate to better long-horizon returns, requiring tight coupling between the world model and the policy optimizer. 
\end{itemize}

These challenges highlight the limitations of existing solutions. End-to-end reactive RL lacks the persistent structure needed for reasoning under occlusion, while static, precomputed scene graphs quickly become outdated.\\

This paper presents a unified \emph{Perception $\rightarrow$ Knowledge $\rightarrow$ Policy} framework, Knowledge-Guided Massively Multi-task Model-based Policy Optimization (KG-M3PO) for multi-task manipulation under partial observability. KG-M3PO augments M3PO with a dynamically updated 3D scene graph whose typed edges (spatial, containment, affordance) are refreshed online via a \emph{dynamic-relation} mechanism. A graph neural encoder produces a compact knowledge embedding that is fused with visual and proprioceptive features; crucially, the encoder is trained \emph{end-to-end through the RL objective}, so relational representations are shaped directly by control performance. \\

\emph{(A) Fully Observable} (camera-only vs.\ camera+KG),
\emph{(B) Multi-Task} (camera-only vs.\ camera+KG), and
\emph{(C) Partially Observable} (camera-only vs.\ camera+KG), featuring tasks like retrieving objects from behind walls or inside closed containers.\\

Our method is benchmarked against strong baselines including PPO~\cite{ppo}, SAC~\cite{sac}, DreamerV3~\cite{dreamerv3}, TD-MPC2~\cite{tdmpc2} and M3PO ~\cite{m3po}. Our results demonstrate that: (1) in fully observable settings, adding the knowledge graph improves sample efficiency for RL agents; (2) in partially observable settings, the dynamic graph is essential for solving tasks that are otherwise impossible from camera-only inputs; and (3) the proposed framework is effective and generalizable. \\

Our principal contributions are:
\begin{itemize}
    \item A unified architecture KG-M3PO that integrates a \emph{dynamically updated scene graph} with the M3PO agent for manipulation under partial observability.
    \item A novel \emph{online relation-update mechanism} that maintains temporally consistent spatial, containment, and affordance relations from a stream of egocentric observations.
    \item A flexible \emph{knowledge-conditioned policy interface} that uses graph queries to provide goal and contextual signals, enabling effective multi-task policy learning.
    \item A comprehensive benchmark and analysis demonstrating that our framework can be applied to train successful multi-task agents in a photorealistic environment with curriculum learning \& domain randomization for robust policy and high vectorization for faster training.
\end{itemize}

\section{Related Work}

\subsection{RL for Manipulation Tasks}
\label{subsec:rl_manipulation}

Modern reinforcement learning (RL) for robot manipulation spans simulation benchmarks, algorithmic advances for sample efficiency and stability, and real-robot evaluations. Standardized suites such as \emph{robosuite}~\cite{zhu2020robosuite}, Meta-World~\cite{yu2020metaworld}, and ManiSkill2~\cite{gu2023maniskill2} provide diverse pick-place, tool-use, and articulated-object tasks with unified APIs, enabling controlled comparisons and large-scale training.

On the algorithmic side, model-based RL has become increasingly competitive for visuomotor control: DreamerV3~\cite{dreamerv3} learns world models that plan via imagination, while TD-MPC2~\cite{tdmpc2} couples latent world models with MPC-style updates and scales to many tasks and embodiments as demonstrated in \cite{leveragingmtrl}. Representation choices remain important for visuo-motor policies; world-model–based visuomotor encoders such as MoDem-V2 and methods that finetune offline world models in the real world improve data efficiency and enable contact-rich real-robot skills~\cite{lancaster2024modemv2,feng2023fowm}.

\subsection{3D scene graphs, open-vocabulary grounding, and LLM reasoning}
3D scene graphs (3DSGs) represent environments as graphs where nodes are objects and edges are their spatial/semantic relationships \cite{armeni20193d}. Early 3DSG approaches \cite{wang2023scene, zhang2021exploiting, koch2024sgrec3d} were closed-vocabulary, limited to fixed object and relation sets. Recent works \cite{gu2024conceptgraphs, koch2024open3dsg, chen2024clip, werby2024hierarchical} have generalized this to open-vocabulary graphs. For example, ConceptGraphs \cite{gu2024conceptgraphs} builds a 3D scene graph by fusing class-agnostic 2D segmentation with CLIP/DINO features and then labeling each object with free-text captions from a vision-language model (LLaVA \cite{liu2023visual}) and relationships inferred by an LLM. Similarly, BBQ \cite{linok2025barequeriesopenvocabularyobject} constructs a 3D scene graph with both metric (distance) and semantic edges and uses an LLM to handle complex, relational queries.

A key challenge in 3DSGs is predicting the edge labels (relationships) between objects, especially under long-tail distributions. VL-SAT \cite{wang2023vl} addresses this by leveraging visual-linguistic semantics during training. VL-SAT trains a powerful multi-modal “oracle” that uses CLIP \cite{radford2021learning} and language features to learn structural priors for object relations, and then distills this knowledge to purely 3D models. In effect, the oracle learns to encode geometric and semantic cues together, and its guidance helps the 3D model better discriminate rare or ambiguous relations.

Together, these advances have rendered 3D scene graphs practical building blocks for language-grounded robotics. Hierarchical open-vocabulary representations such as HOV-SG and Point2Graph compress dense geometric maps into multilevel floor/room/object abstractions that scale to large, multi-story environments \cite{werby2024hierarchical, yin2024sg}. Built on these representations, online prompting and hierarchical chain-of-thought traversal SG-Nav enable LLMs to reason over graph structure to infer object locations and navigation goals, yielding explainable zero-shot navigation \cite{yin2024sg}. Complementarily, dynamic 3DSG DovSG model temporal change and support long-horizon, change-aware mobile manipulation \cite{yan2025dynamic}. Collectively, these methods improve scalability, explainability, and zero-shot generalization in vision–language robotics. This work builds upon these advances by integrating a dynamically updated, open-vocabulary 3D scene graph for robotic manipulation.

\subsection{Knowledge-guided manipulation}
\label{subsec:kg_pomdps}
We model manipulation as a partially observable Markov decision process (POMDP), where the agent must maintain latent state under occlusions and distribution shifts. A growing body of work demonstrates that explicit graph-structured knowledge (e.g., semantic maps or scene graphs) serves as a compact belief state, improving exploration, long-horizon reasoning, and credit assignment. For instance, semantic maps enable hierarchical RL for mobile manipulation (HIMOS)~\cite{aditya-himos-2023} and open-vocabulary grounding (OVMM)~\cite{aditya-ovmm-2023}. Dynamic scene-graph methods further advance this: MoMa-LLM~\cite{moma-llm-2024} grounds instructions in updated scene graphs, RoboEXP~\cite{roboexp-2024} builds action-conditioned graphs for exploration, and GRID~\cite{grid-2023} fuses instruction, scene, and robot graphs for task planning.

\section{Problem Setup}
\subsection{Partially Observable MDP formalization}
We model each task as a \textit{partially observable} MDP (POMDP) \(\mathcal{M} = \langle S, A, P, R, \gamma \rangle\). The true state \( s \in S \) (which is not fully observed by the agent) includes the full physical configuration: the robot’s joint angles, the object poses, and any relevant environmental variables (e.g., drawer open percentage). The agent receives an observation \( o(s) \) which, in our case, is either (i) an RGB image from the wrist camera (used by the camera-only \textbf{M3PO} baseline), or (ii) the same RGB image augmented with a learned knowledge vector from the online scene graph (used by \textbf{KG-M3PO}). The \textbf{transition function} \( P(s_{t+1} \mid s_t, a_t) \) is governed by the physics simulator – effectively a deterministic dynamical system given the action (since we use a fixed simulation seed for determinism, ignoring minor sensor noise). In practice, \( P \) is unknown to the agent and we treat the simulator as a black box environment step function. We discount future rewards with factor \(\gamma=0.99\) in all tasks.

\subsection{Reward Design}
Each task uses a shaped reward to guide the agent toward success while avoiding bad behaviors. We emphasize \textbf{dense rewards for motion guidance} (like distances) and \textbf{sparse rewards for goal completion} (like successfully lifting or opening), following common practice in robotic RL. Terminal conditions define success or failure: \textit{success} yields an episode completion with a high terminal reward, while \textit{failure} (e.g., dropping an object or exceeding time limit) ends the episode with zero or negative reward.

\textbf{General Reward Formula}
For any task $i$ with goal $g_i$, state $s_t$, action $a_t$, joints $q_t$, and (optional) knowledge graph $\mathcal{K}_t$, we use:
\begin{equation}
\label{eq:general-reward}
\begin{aligned}
r_t^{(i)} = &\underbrace{\alpha_{\mathrm{succ}}^{(i)}\,\mathbb{I}\!\left[\mathrm{success}_i(s_{t+1};g_i)\right]}_{\text{terminal success}} \\
&+ \underbrace{\gamma_{\Phi}\!\left(\Phi_i(s_{t+1};g_i,\mathcal{K}_{t+1})-\Phi_i(s_t;g_i,\mathcal{K}_{t})\right)}_{\text{potential-based shaping}} \\
&+ \underbrace{\boldsymbol{\beta}^{(i)\!\top}\Delta\boldsymbol{\psi}\!\left(s_t,a_t,s_{t+1},\mathcal{K}_t\right)}_{\text{event/relational increments}} \\
&- \underbrace{\lambda_a \|a_t\|_2^2 - \lambda_j \|q_{t+1}-q_t\|_2^2}_{\text{action/joint penalties}} \\
&- \underbrace{\lambda_c \,\mathbb{I}[\mathrm{collision}(s_{t+1})] - \lambda_t}_{\text{collision/time penalties}},
\end{aligned}
\end{equation}
\noindent\textit{where:} $\Phi_i$ is a scalar potential function; $\Delta\boldsymbol{\psi}$ represents task-specific incremental signals; and $\alpha_{\mathrm{succ}}^{(i)},\gamma_{\Phi},\boldsymbol{\beta}^{(i)},\lambda_a,\lambda_j,\lambda_c,\lambda_t$ are tunable weights.

\subsection{Observation models: M3PO vs.\ KG-M3PO}
\label{subsec:observation_models}

We evaluate two variants that share the same M3PO control backbone but differ in what is provided as input to the policy:

\begin{enumerate}
    \item \textbf{M3PO (camera-only baseline):} $o_t = I_t \in \mathbb{R}^{H \times W \times 3}$ (wrist camera, $H=W=64$).
    \item \textbf{KG-M3PO (camera + knowledge graph):} $o_t=(I_t,k_t)$ with $I_t\in\mathbb{R}^{64\times64\times3}$ and $k_t=\mathrm{GNN}_{\varphi}(K_t)\in\mathbb{R}^{32}$. The scene graph $K_t$ is updated \emph{online} every step from the current egocentric observation; the GNN runs each step to produce $k_t$, and its parameters $\varphi$ are trained end-to-end through the RL loss (Sec.~\ref{sec:e2e-arch}).

\end{enumerate}

The precise implementation details for both inputs, including image resolution and the feature extraction process for $k_t$, are provided in the Experimental Setup (Section~\ref{subsec:observation_space_impl}).



\section{Unified Framework: Perception $\rightarrow$ Knowledge $\rightarrow$ Policy}

\subsection{Knowledge Graph}
\label{subsec: knowledge-graph}
\begin{figure}[t]
    \centering
    
    \begin{minipage}[t]{0.49\columnwidth}
        \centering
        \includegraphics[width=\linewidth]{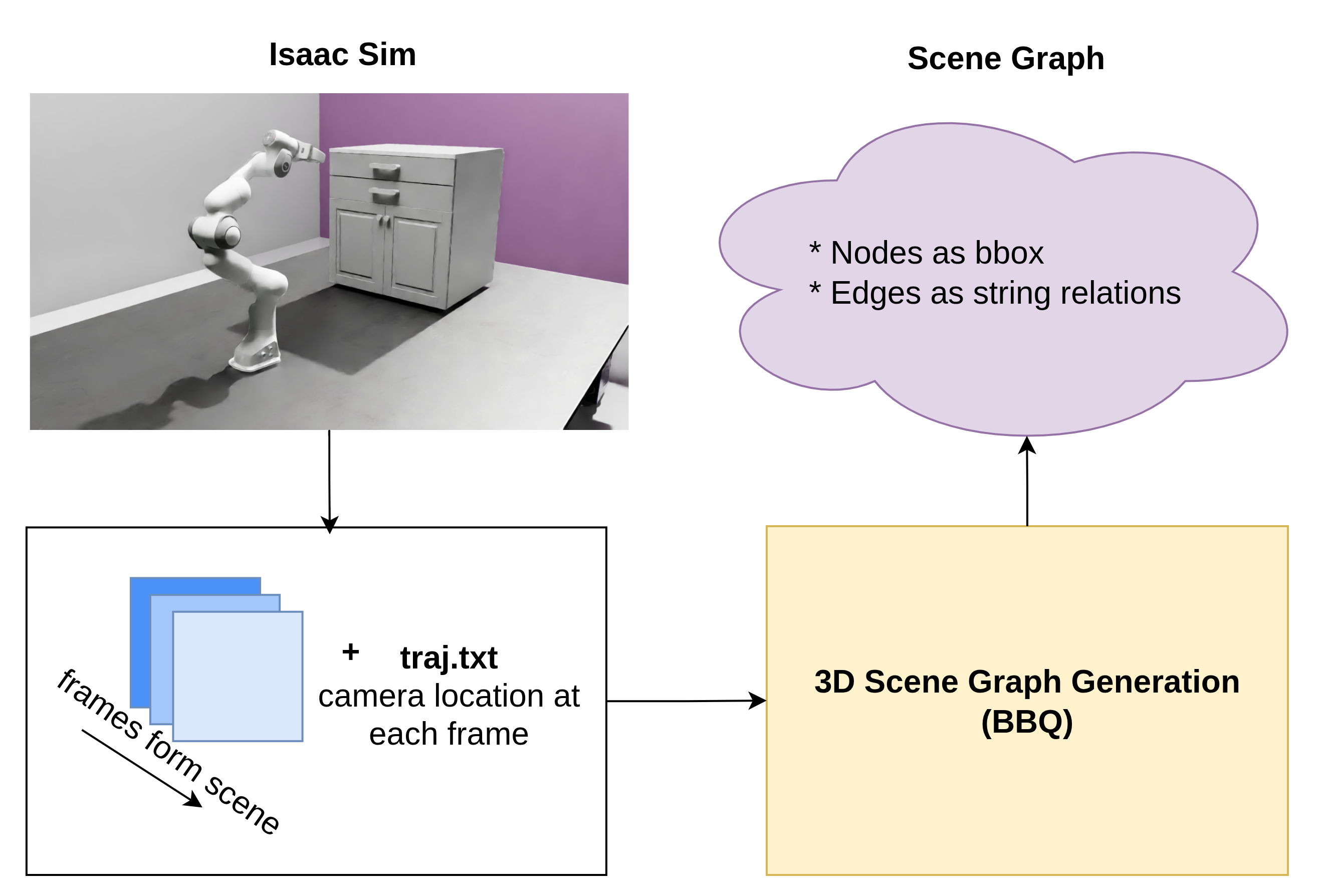}
        \caption{Scene Graph generation pipeline.}
        \label{fig:data_collect}
    \end{minipage}
    \hfill
    \begin{minipage}[t]{0.49\columnwidth}
        \centering
        \includegraphics[width=\linewidth]{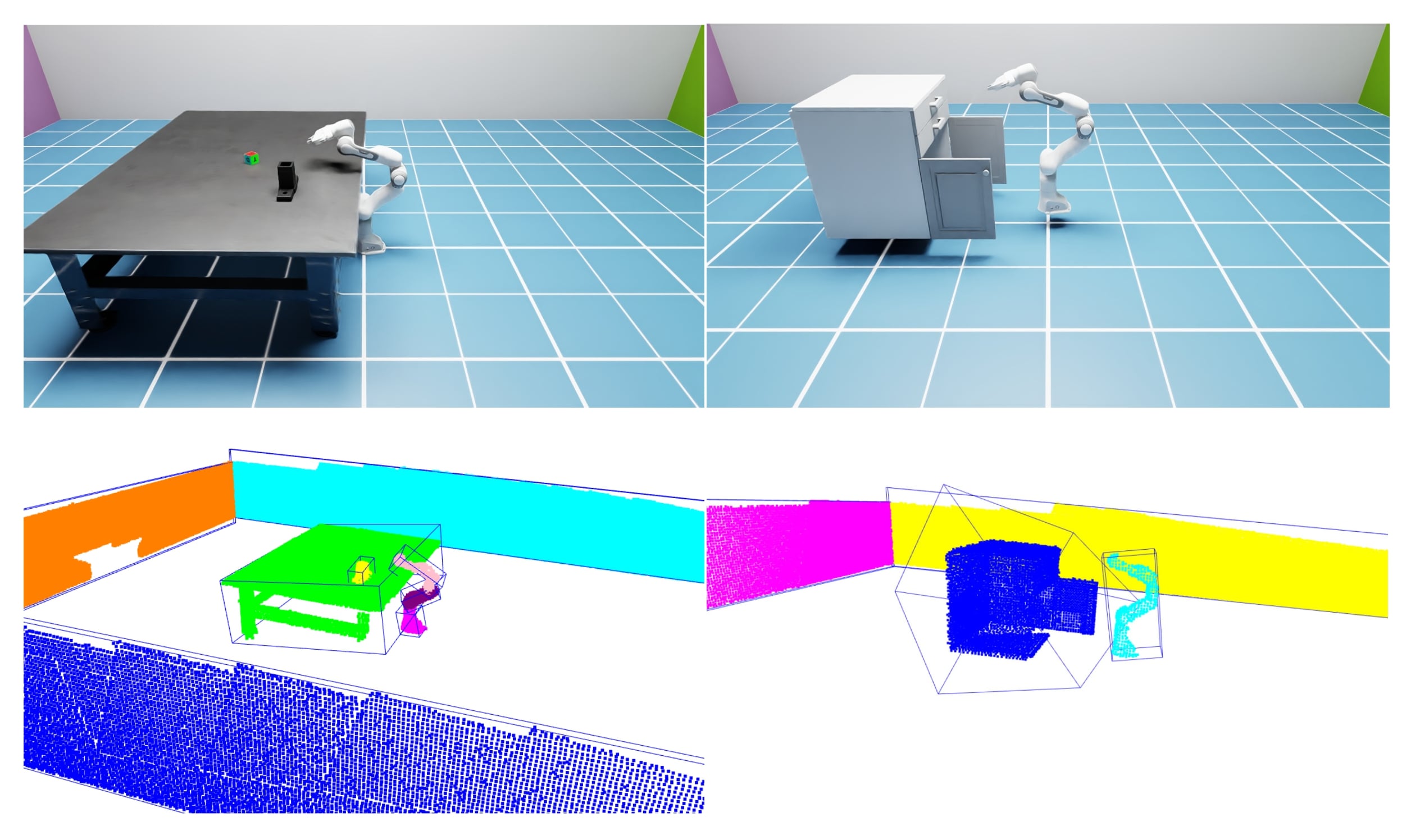}
        \caption{Example BBQ output with Franka, table, and cabinet.}
        \label{fig:pcd}
    \end{minipage}

\end{figure}

We construct scene graphs using BBQ~\cite{linok2025barequeriesopenvocabularyobject}. Compared to similar approaches, BBQ~\cite{linok2025barequeriesopenvocabularyobject} demonstrated high efficiency in generating a scene graph from a sequence of the robot's sensory data, as well as high-quality object grounding, which is particularly important for solving object manipulation tasks. Scenes are created in Isaac Sim; we collect RGB--D sequences along predefined camera trajectories and process the recordings with BBQ. Each graph node encodes an object's 3D bounding-box center, bounding-box extent, and a 512-dimensional CLIP~\cite{radford2021learning} embedding. To improve robustness, we fine-tune the generated graphs using simulator ground truth (object identities and 6D poses).
Our data-collection pipeline is illustrated in Fig.~\ref{fig:data_collect}, and an example BBQ output is shown in Fig.~\ref{fig:pcd}.

The graph also encodes pairwise spatial relations derived from oriented bounding boxes and simple contact/visibility tests. In our tasks, the most informative relations are \emph{in front of}, \emph{behind}, \emph{on}, and \emph{under}; we compute and maintain these edges online and expose them to the policy.

A lightweight detector flags moved objects from the egocentric stream. Every $n=10$ steps, we refresh affected subgraphs: update CLIP embeddings and recompute local spatial relations. This partial update ensures low overhead.


\subsection{End-to-End Architecture}
\label{sec:e2e-arch}

\textbf{Fusion for control.}  
At step $t$ we form three embeddings and a fused policy state:
\begin{align}
z_t^{\mathrm{img}} &= f_{\psi}(I_t) \in \mathbb{R}^{d_I}, \\
q_t &= h_{\phi}(K_t) \in \mathbb{R}^{d_K}, \\
u_t &= e(g_t) \in \mathbb{R}^{d_T}, \\[2pt]
s_t &= \big[\,W_I z_t^{\mathrm{img}} \;\Vert\;
        W_K q_t \;\Vert\;
        W_T u_t \;\Vert\;
        x^{\mathrm{prop}}_t \big] \in \mathbb{R}^{d_s}.
\label{eq:fused-state}
\end{align}

\textbf{Control heads.}  
We use Gaussian policy and scalar value heads:
\begin{align}
\pi_\theta(a_t\!\mid\!s_t)
 &= \mathcal{N}\!\left(\mu_\theta(s_t),
    \mathrm{Diag}\!\big(\sigma^2_\theta(s_t)\big)\right), \\
V_\xi(s_t) &\in \mathbb{R}.
\label{eq:gauss}
\end{align}

The per-update control loss combines policy, value, and entropy:
\begin{align}
\mathcal{J}_{\mathrm{ctrl}}
&= -\mathcal{L}_{\mathrm{M3PO}}
   + \beta_V\,\mathbb{E}_t\!\big(V_\xi(s_t)-\hat R_t\big)^2 \nonumber\\
&\quad - \beta_H\,\mathbb{E}_t\!\big[\mathcal{H}(\pi_\theta(\cdot\mid s_t))\big].
\label{eq:ctrl}
\end{align}

\textbf{GNN directly inside the loss.}  
Unlike detached graph encoders, our architecture makes the GNN part of the
\emph{RL loss computation itself}.  
Specifically, the knowledge encoder is a GNN with parameters $\varphi$:
\[
q_t = \mathrm{GNN}_\varphi(K_t),
\]
and because $q_t$ flows into $s_t$ (\eqref{eq:fused-state}), the policy
gradient backpropagates through the GNN.  
For M3PO, the update to $\varphi$ is
\begin{align}
\nabla_{\varphi}\,\mathcal{J}_{\mathrm{ctrl}}
&= -\,\mathbb{E}_t\!\Big[
  \underbrace{\nabla_{s_t}\log \pi_\theta(a_t\!\mid\!s_t)\,\hat A_t}_{\text{policy}}
  \tfrac{\partial s_t}{\partial q_t}
  \tfrac{\partial q_t}{\partial \varphi}
  \Big] \nonumber\\
&\quad + \beta_V\,\mathbb{E}_t\!\Big[
  \nabla_{s_t}(V_\xi(s_t)\!-\!\hat R_t)^2
  \tfrac{\partial s_t}{\partial q_t}
  \tfrac{\partial q_t}{\partial \varphi}\Big].
\label{eq:gnn-grad}
\end{align}

Eq.~\ref{eq:gnn-grad} shows that the GNN is \emph{trained end-to-end by the M3PO loss itself}, not by auxiliary supervision. The same integration principle holds for any policy-gradient or actor–critic algorithm (\emph{e.g.}, PPO, SAC, TD-MPC2): once $q_t$ is fused into $s_t$, the chosen control loss automatically propagates gradients into the GNN.

\section{Benchmarking Environment and Baseline Algorithms}

\subsection{Environment}

\subsubsection{Benchmarking environment}
Our benchmarking environment (see Fig.~\ref{fig:train-pipeline}) is built on \textbf{NVIDIA Isaac Lab}, based on \cite{isaaclab} atop \textbf{Isaac Sim}, using \textbf{GPU-accelerated PhysX 5} for dynamics and multi-sensor \textbf{RTX} rendering for photorealistic, physics-consistent observations. For vision-in-the-loop RL at scale, we use Isaac Lab's \textbf{tiled rendering} API, which vectorizes camera readout by rendering a single stitched image from multiple per-environment cameras, substantially reducing overhead for large batched rollouts. Isaac Lab provides vectorized RL environments and training utilities to run many environments in parallel within one process. In our experiments we execute \textbf{NumEnvs = 1024} parallel environments on a single workstation equipped with an \textbf{NVIDIA RTX 5070\,Ti (16\,GB VRAM)} and \textbf{64\,GB RAM}. 

\textbf{Curriculum learning.} IsaacLab exposes a \emph{CurriculumManager} that applies configurable \emph{curriculum terms} during training to progressively harden the task (e.g., tightening success tolerances and modifying environment parameters). 

\textbf{Domain randomization.} Robustness is improved through an \emph{EventManager} that triggers randomization at startup, reset, or fixed intervals (e.g., masses, friction, external pushes), combined with Isaac Sim’s Replicator to vary appearance and sensor parameters (textures, lighting, camera intrinsics/noise) and with “on-the-fly” dynamics randomization.

\subsubsection{Robot platforms}
We evaluate on two collaborative manipulators: the \textbf{Franka Emika Panda} (7-DoF) with the native parallel-jaw Franka gripper, and the \textbf{Universal Robots UR5} (6-DoF) with a \textbf{Robotiq 2F-series} parallel-jaw gripper. 

\subsubsection{Observation Space}
\label{subsec:observation_space_impl}
We implement the two observation models defined in Section~\ref{subsec:observation_models} as follows:

\begin{itemize}
    \item \textbf{M3PO (camera-only):} The RGB image $I_t$ is rendered at a resolution of $64 \times 64$ pixels from a wrist-mounted pinhole camera (Isaac Lab \texttt{TiledCameraCfg}).
    \item \textbf{KG-M3PO (camera + KG):} The image $I_t$ is $64\times64$. The knowledge vector $k_t\in\mathbb{R}^{32}$ is computed \emph{every control step} by a graph neural network $k_t=\mathrm{GNN}_{\varphi}(K_t)$ from the current scene graph $K_t$; $\varphi$ receives gradients from $\mathcal{J}_{\mathrm{ctrl}}$ (Eq.~\ref{eq:gnn-grad}).
\end{itemize}

\subsubsection{Action space:}
The agent issues continuous end-effector \emph{delta pose} commands (Cartesian SE(3) twist: translation and rotation increments) plus a scalar gripper command; a whole-body controller maps these to joint targets per manipulator. Episodes run in discrete control steps (e.g., \(\sim\!60\)~Hz) and terminate on success, failure, or timeout. Actions are normalized to \([-1,1]\) per dimension, with optional smoothness penalties in the reward shaping.

\subsection{Baselines}
We benchmark six widely used RL algorithms across \emph{single-task} and \emph{multi-task} settings: on-policy \textbf{PPO}, off-policy \textbf{SAC}, model-based \textbf{DreamerV3} and \textbf{TD-MPC2}, distributed actor–learner \textbf{IMPALA}~\cite{impala}, and hybrid model-based/on-policy \textbf{M3PO}.

\textbf{Single-task.} Each method is trained separately on each manipulation task under two input regimes: \emph{camera-only} and \emph{camera + knowledge graph}. For the M3PO backbone, we refer to these two variants as \textbf{M3PO} (camera-only) and \textbf{KG-M3PO} (camera+KG). For all other baselines, we keep the algorithm name (e.g., PPO, SAC) and specify the input regime in the corresponding plots/tables.

\textbf{Multi-task.} A \emph{single} policy/value (or world model + policy) is trained jointly on all tasks with a one-hot \emph{task ID} concatenated to observations (denoted MT-PPO/MT-SAC; M3PO, TD-MPC2 and IMPALA are inherently multi-task). Unless noted in ablations, multi-task runs use the \emph{camera + knowledge graph} input regime (the partially observable tasks are otherwise unsolvable from pixels alone); we also report a camera-only variant to isolate the impact of knowledge inputs.

\subsection{Evaluation}
\label{subsec:evaluation}

To ensure fair and interpretable comparison across diverse tasks and reward scales, we adopt a normalized scoring metric standard in reinforcement learning benchmarks.
For each task \(i\), let
\[
\begin{aligned}
R_i               &= \text{agent’s average return},\\
R_i^{\mathrm{rand}} &= \text{random‐policy baseline},\\
R_i^{\mathrm{exp}}  &= \text{expert or single‐task upper bound}.
\end{aligned}
\]

We define the normalized score \(s_i\) by
\[
\begin{aligned}
s_i 
&= \mathrm{clip}\!\Bigl(\,
      \frac{R_i - R_i^{\mathrm{rand}}}
           {R_i^{\mathrm{exp}} - R_i^{\mathrm{rand}}},
      \,0,\,1
    \Bigr)
    \times 1000\,. 
\end{aligned}
\]

\begin{enumerate}
  \item \textbf{Anchoring to chance performance.}  
    Raw returns \(R_i\) can vary widely in scale (and even sign) across tasks.  
    By subtracting the random‐policy return \(R_i^{\mathrm{rand}}\), we shift every task so that “zero” corresponds exactly to what one would achieve by taking actions uniformly at random.  
    Clamping the fraction below zero ensures no negative scores, and scaling by 1000 makes “0” an intuitive lower bound meaning “no better than chance.”

  \item \textbf{Cross‐task comparability.}  
    Different tasks often have very different reward scales: a return of 50 might be excellent on one task but poor on another.  
    Dividing by \(\bigl(R_i^{\mathrm{exp}} - R_i^{\mathrm{rand}}\bigr)\) maps the expert‐level performance to “1,” and multiplying by 1000 maps it to “1000”.
\end{enumerate}

\section{Experiments}

\begin{figure}[t]
    \centering
    \begin{subfigure}[b]{0.32\columnwidth}
        \centering
        \includegraphics[width=\linewidth]{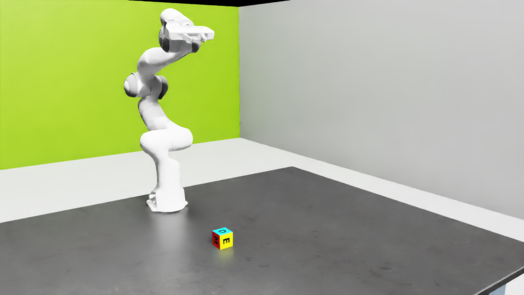}
        \caption{Pick-Cube}
        \label{fig:snap-pick}
    \end{subfigure}\hfill
    \begin{subfigure}[b]{0.32\columnwidth}
        \centering
        \includegraphics[width=\linewidth]{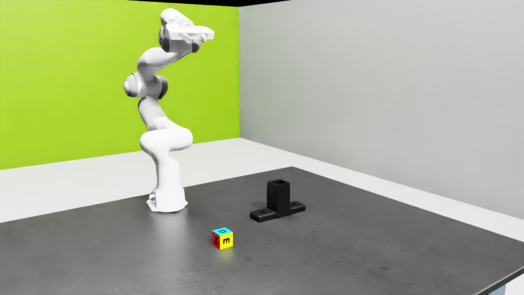}
        \caption{Pick-Place}
        \label{fig:snap-pickplace}
    \end{subfigure}
    \begin{subfigure}[b]{0.32\columnwidth}
        \centering
        \includegraphics[width=\linewidth]{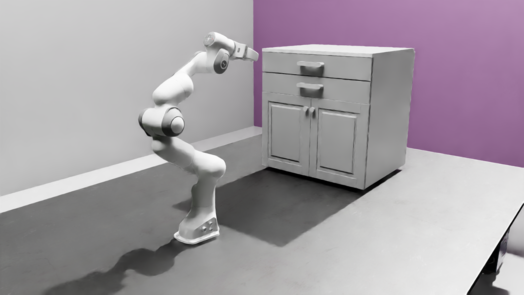}
        \caption{Open-Cabinet}
        \label{fig:snap-cabinet}
    \end{subfigure}\hfill

    \caption{\textbf{Benchmark snapshots.} Franka scenes for three representative tasks used in our study. 
    Identical task variants and environments are also implemented for \textbf{UR5} (not shown).}
    \label{fig:env-snapshots}
\end{figure}

\subsection{Tasks}

\begin{table}[t]
\centering
\caption{Benchmark tasks.}
\label{tab:tasks-small}
\scriptsize
\setlength{\tabcolsep}{3pt}
\renewcommand{\arraystretch}{1.05}

\begin{tabular}{l c p{4.2cm}}
\toprule
\textbf{Task} & \textbf{Obs.} & \textbf{Description} \\
\midrule
Pick-Cube & FO & Pick a colored cube from the table. \\

Pick-Place & FO & Pick a cube and place it in the square peg. \\

Open-Cabinet & FO & Grasp the handle and open the drawer. \\

PO Pick & PO & Pick a cube initially occluded by a wall. \\

PO Pick-Place & PO & Remove occluder (cone), then pick and place the cube. \\
\bottomrule
\end{tabular}

\smallskip
\footnotesize
FO: fully observable. PO: partially observable.
\end{table}

\begin{figure}[t]
    \centering
    \begin{subfigure}[b]{0.32\columnwidth}
        \centering
        \includegraphics[width=\linewidth]{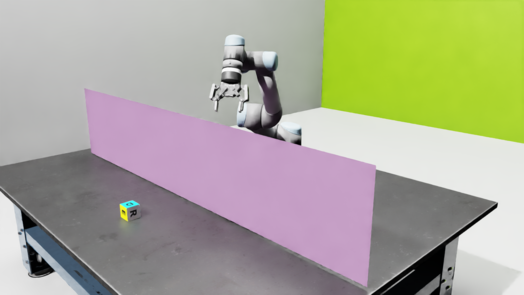}
        \caption{PO Pick: target hidden behind a wall (start).}
        \label{fig:po-a}
    \end{subfigure}\hfill
    \begin{subfigure}[b]{0.32\columnwidth}
        \centering
        \includegraphics[width=\linewidth]{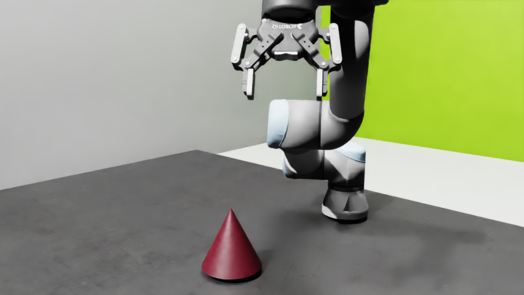}
        \caption{PO Pick-Place: occluder present (start).}
        \label{fig:po-2a}
    \end{subfigure}\hfill
    \begin{subfigure}[b]{0.32\columnwidth}
        \centering
        \includegraphics[width=\linewidth]{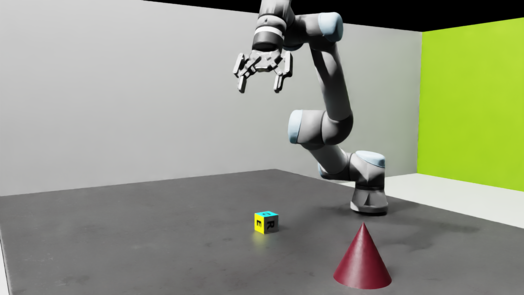}
        \caption{PO Pick-Place: occluder moved by the arm.}
        \label{fig:po-2b}
    \end{subfigure}

    \caption{\textbf{Partially observable tasks.} We show UR5 examples for two PO scenarios:
    (a) picking an object initially hidden by a wall; (b–c) a two-stage pick–place where the agent must first remove an occluder and then retrieve the target. \emph{Identical task variants are implemented for both \textbf{Franka} and \textbf{UR5}}.}
    \label{fig:po-snapshots}
\end{figure}

We evaluate on five manipulation tasks spanning \emph{fully observable} (see Fig.~\ref{fig:env-snapshots}) and \emph{partially observable} regimes (see Fig.~\ref{fig:po-snapshots}). All tasks have a horizon length of 24 except the \emph{Partially-Observable Pick-Place} task, which requires 48 steps due to its multi-stage nature. 

\begin{figure*}[t]
  \centering
  \setlength{\tabcolsep}{0pt}
  \begin{subfigure}{0.29\textwidth}
    \includegraphics[width=\linewidth]{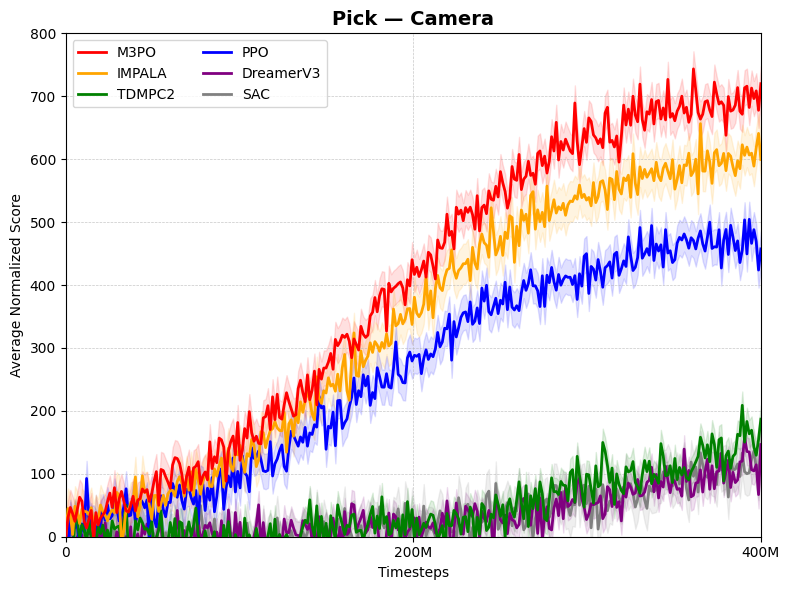}
    \caption{Pick (camera-only)}
  \end{subfigure}
  \hfill
  \begin{subfigure}{0.29\textwidth}
    \includegraphics[width=\linewidth]{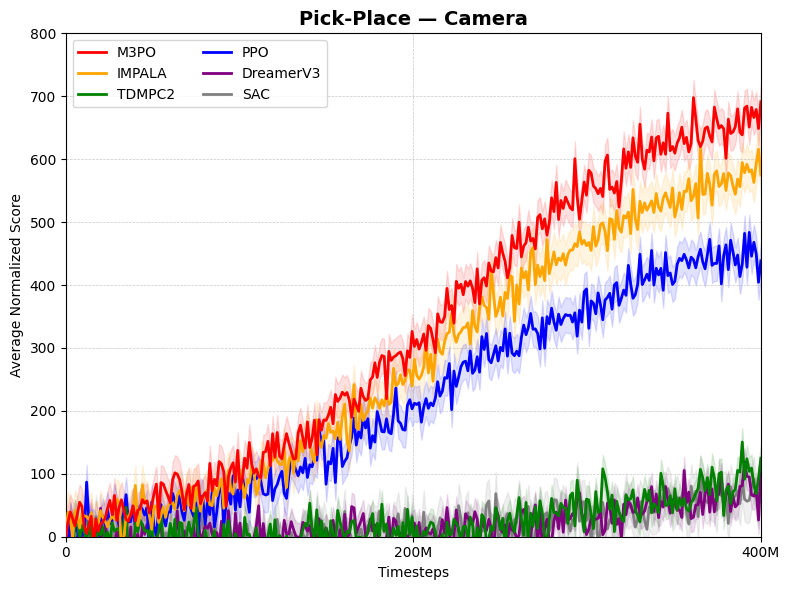}
    \caption{Pick-Place (camera-only)}
  \end{subfigure}
  \hfill
  \begin{subfigure}{0.29\textwidth}
    \includegraphics[width=\linewidth]{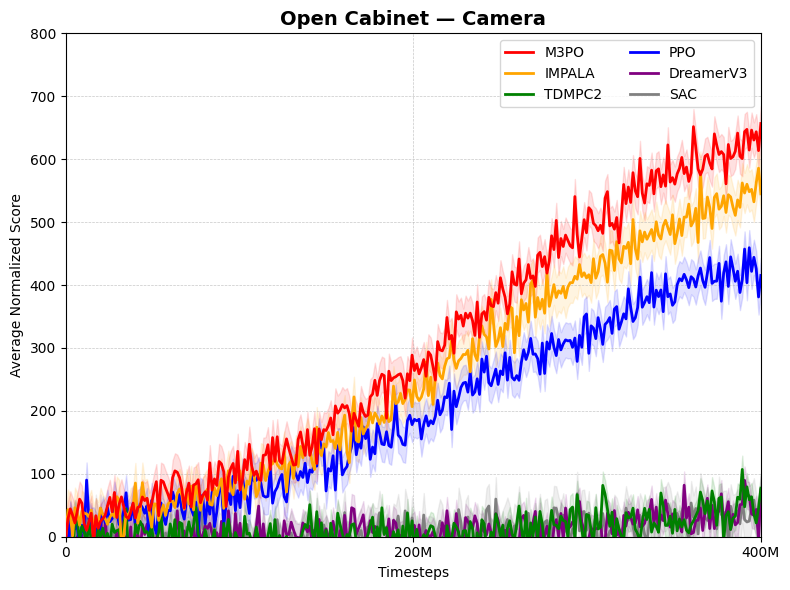}
    \caption{Open-Cabinet (camera-only)}
  \end{subfigure}
  \vspace{0.3em}
  \begin{subfigure}{0.29\textwidth}
    \includegraphics[width=\linewidth]{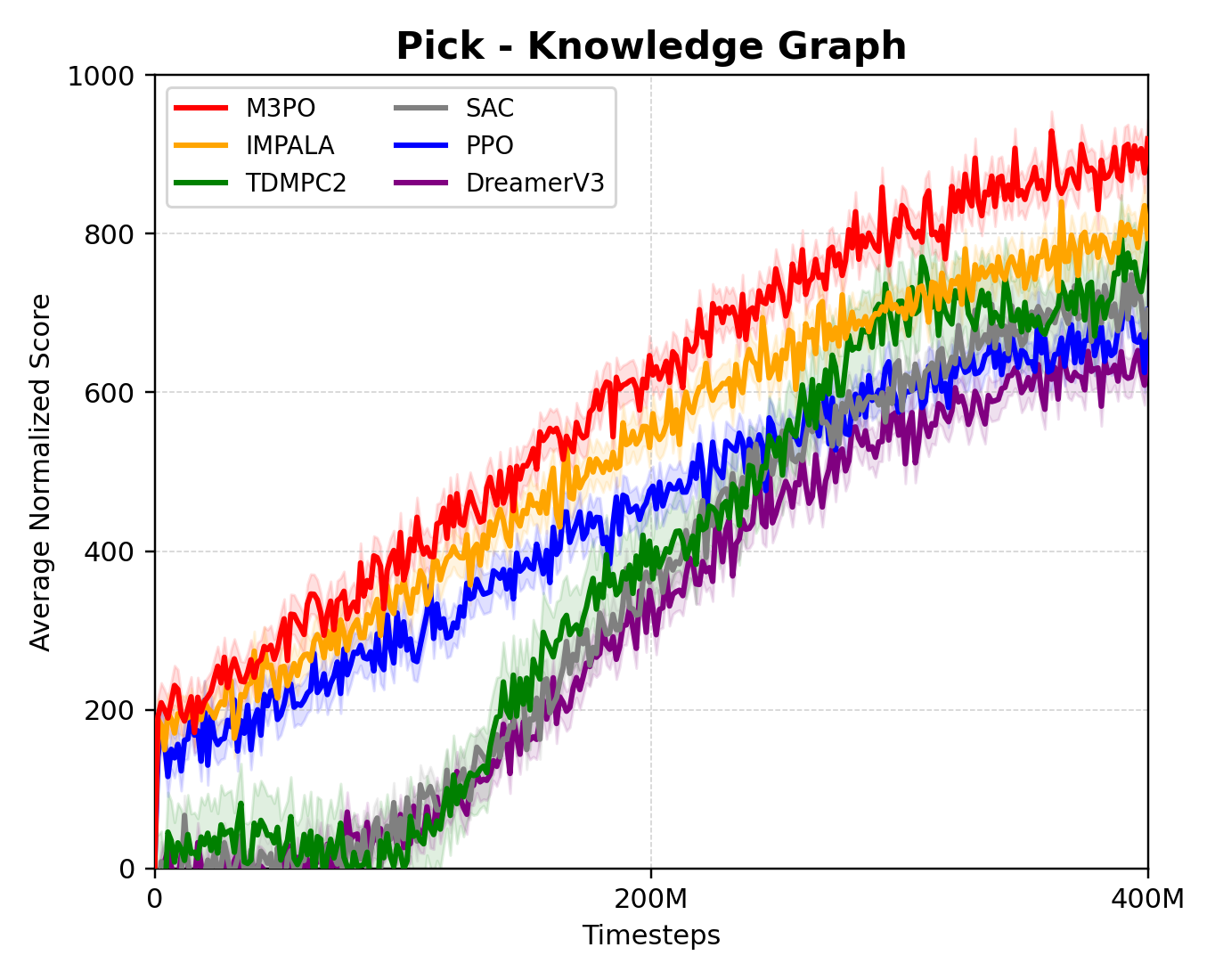}
    \caption{Pick (camera+KG)}
  \end{subfigure}
  \hfill
  \begin{subfigure}{0.29\textwidth}
    \includegraphics[width=\linewidth]{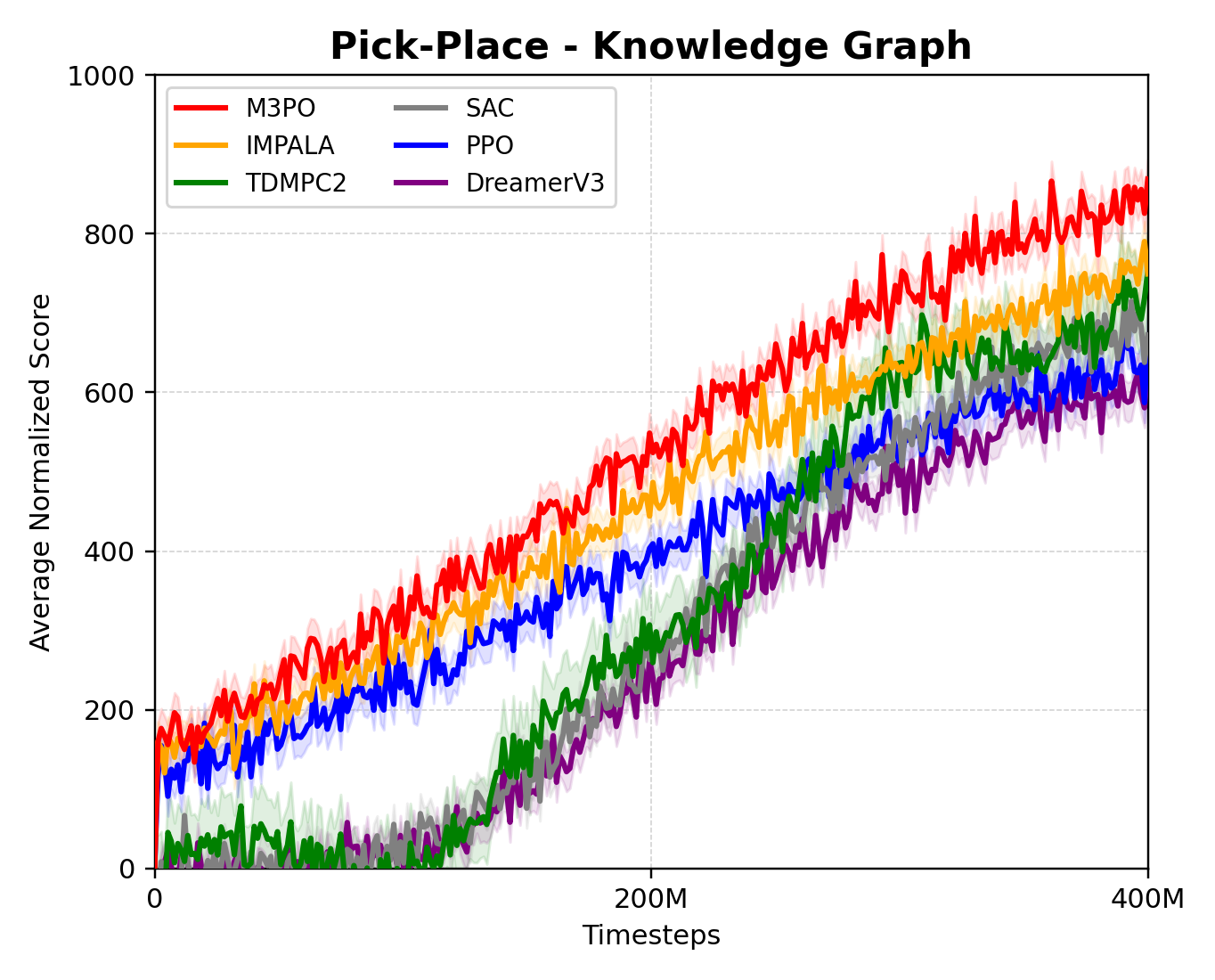}
    \caption{Pick-Place (camera+KG)}
  \end{subfigure}
  \hfill
  \begin{subfigure}{0.29\textwidth}
    \includegraphics[width=\linewidth]{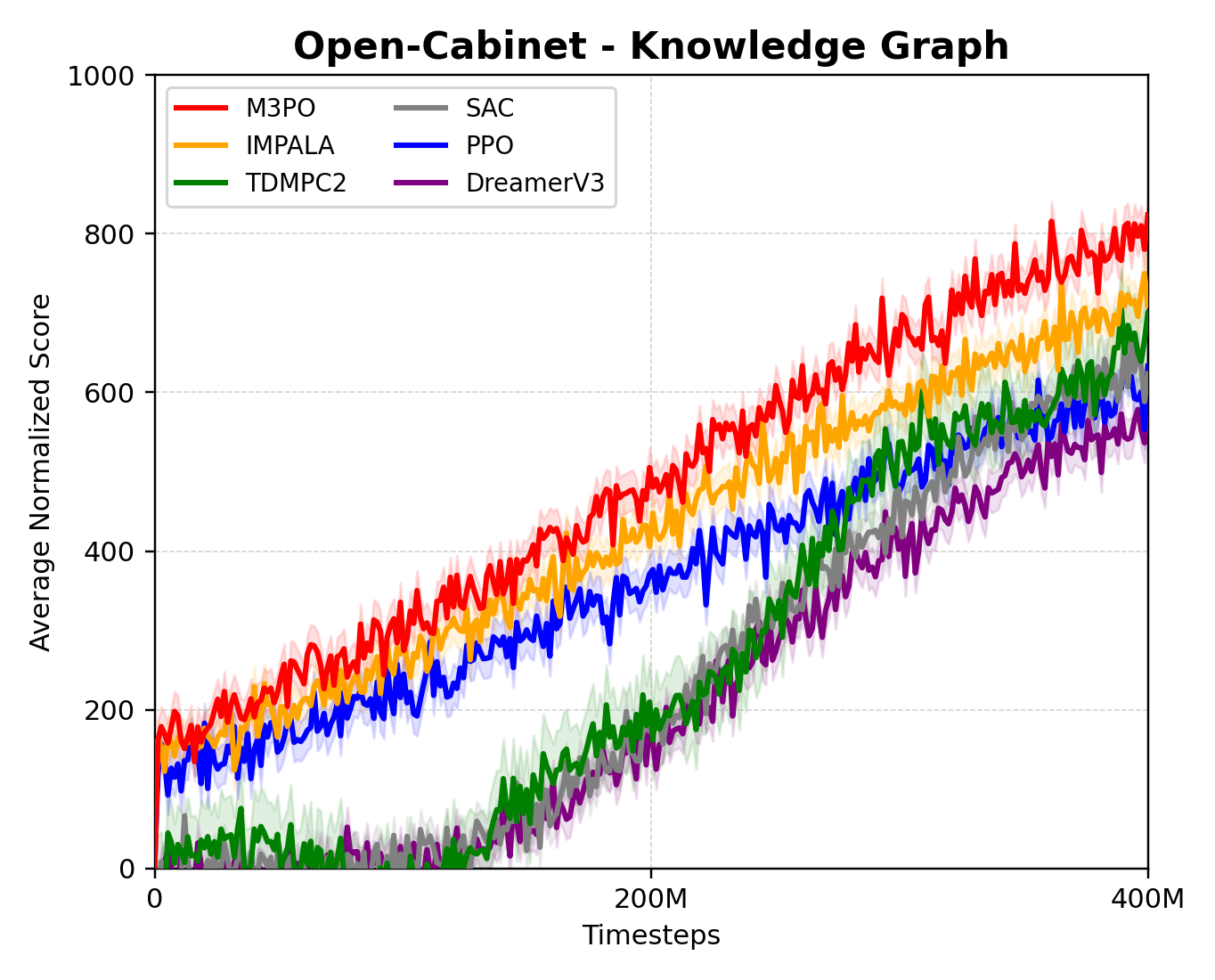}
    \caption{Open-Cabinet (camera+KG)}
  \end{subfigure}
  \caption{\textbf{Single-task results.} Average normalized score vs.\ timesteps. Adding the KG improves sample efficiency and final score. For the M3PO backbone, the camera-only variant corresponds to \textbf{M3PO}, while camera+KG corresponds to \textbf{KG-M3PO}.}
  \label{fig:single-task-clubbed}
\end{figure*}

\begin{table}[t]
\centering
\caption{\textbf{Short- \& long-horizon performance (M3PO vs.\ KG-M3PO).}}
\label{tab:horizon-metrics-small}
\scriptsize
\setlength{\tabcolsep}{2.5pt}
\renewcommand{\arraystretch}{1.05}

\begin{tabular}{l c c c c c}
\toprule
\textbf{Task} & \textbf{Reg.} & \textbf{Succ.} & \textbf{Final} & \textbf{AUC} & \textbf{Steps} \\
\midrule

\multicolumn{6}{l}{\textit{Short horizon ($T{=}24$)}} \\

Pick-Cube & FO & 71 / 87 & 760 / 870 & 410 / 520 & 280 / 120M \\
Pick-Place & FO & 55 / 76 & 670 / 810 & 340 / 460 & 320 / 165M \\
Open-Cabinet & FO & 50 / 72 & 630 / 780 & 300 / 430 & 360 / 190M \\
PO Pick & PO & 4 / 58 & 75 / 640 & 35 / 310 & N/A / 210M \\

\midrule
\textbf{Mean (short)} & & \textbf{45 / 73} & \textbf{534 / 775} & \textbf{271 / 430} & -- \\
\midrule

\multicolumn{6}{l}{\textit{Long horizon ($T{=}48$)}} \\

PO Pick-Place & PO & 2 / 63 & 85 / 670 & 30 / 270 & N/A / 180M \\

\bottomrule
\end{tabular}

\smallskip
\footnotesize
Values are reported as \textit{M3PO (camera-only) / KG-M3PO (camera+KG)}.
FO: fully observable, PO: partially observable.
\end{table}

\subsection{Overall quantitative results and per-task breakdown (Single Task, camera-only vs.\ camera+KG)}
\label{sec:single-task-summary}

We compare performance on short-horizon ($T{=}24$) and long-horizon ($T{=}48$) tasks. Table~\ref{tab:horizon-metrics-small} summarizes comprehensive metrics.

Across all three fully observable tasks, adding the knowledge graph consistently improves sample efficiency while also yielding slightly higher final scores as seen in Fig.~\ref{fig:single-task-clubbed}. In our single-task comparisons, \textbf{M3PO} leads throughout, \textbf{IMPALA} is second, and \textbf{TD-MPC2} generally ranks third (with some instability in camera-only). \textbf{SAC}, \textbf{PPO}, and \textbf{DreamerV3} trail the top trio in both input regimes. The \emph{Pick} task shows only modest gains from KG (as expected for a simple, single-object goal), whereas \emph{Pick-Place} and \emph{Open-Cabinet} benefit more markedly: KG reduces time-to-threshold by directing attention to goal-relevant entities (receptacle/handle) and disambiguating the objective.

\subsection{Multi-Task (camera-only)}
\begin{figure}[t]
  \centering
  \begin{minipage}[t]{0.49\columnwidth}
    \centering
    \includegraphics[width=\linewidth]{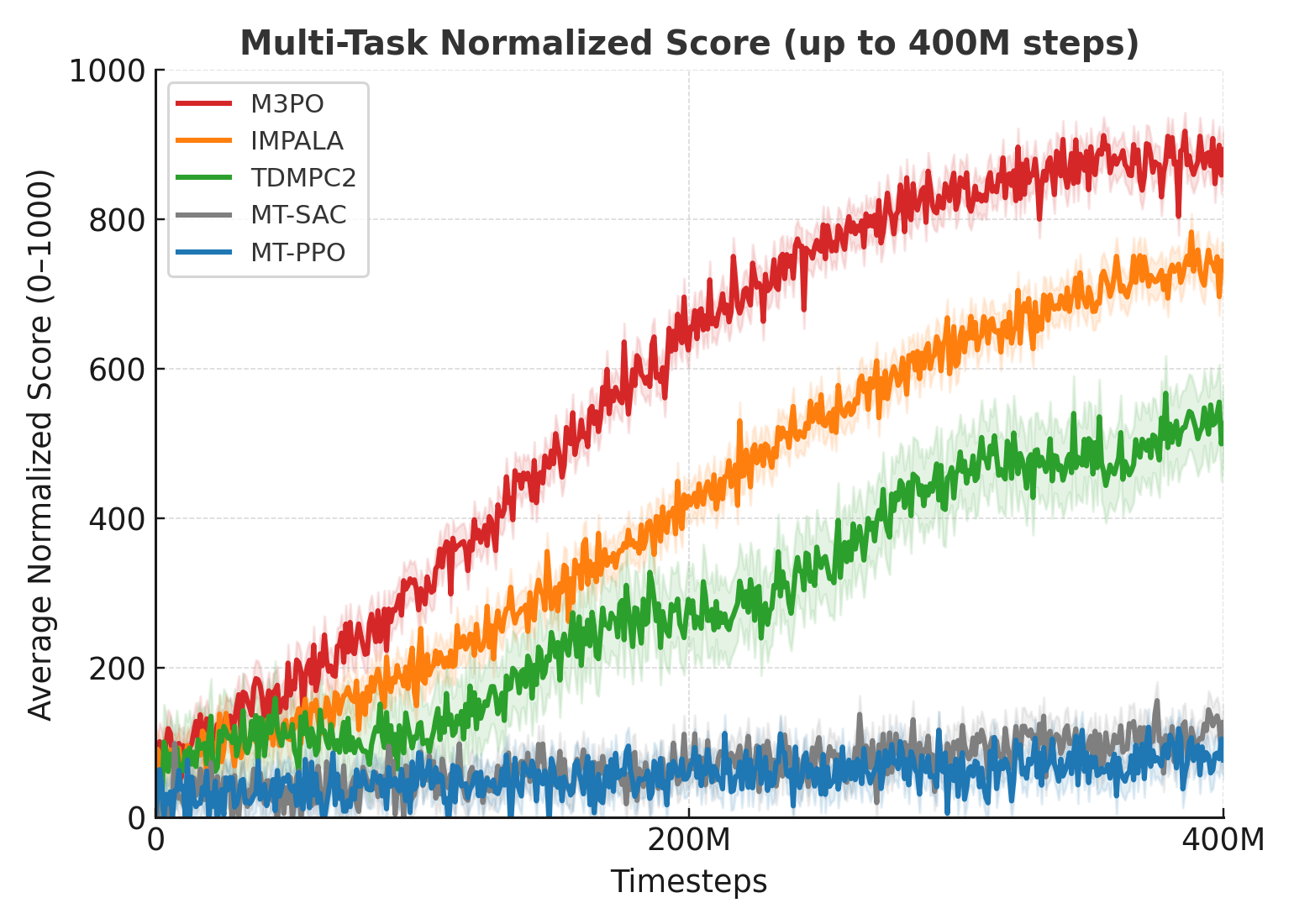}
    \caption{\textbf{Multi-Task (camera-only).} Avg.\ normalized score.}
    \label{fig:mt-camera}
  \end{minipage}
  \hfill
  \begin{minipage}[t]{0.49\columnwidth}
    \centering
    \includegraphics[width=\linewidth]{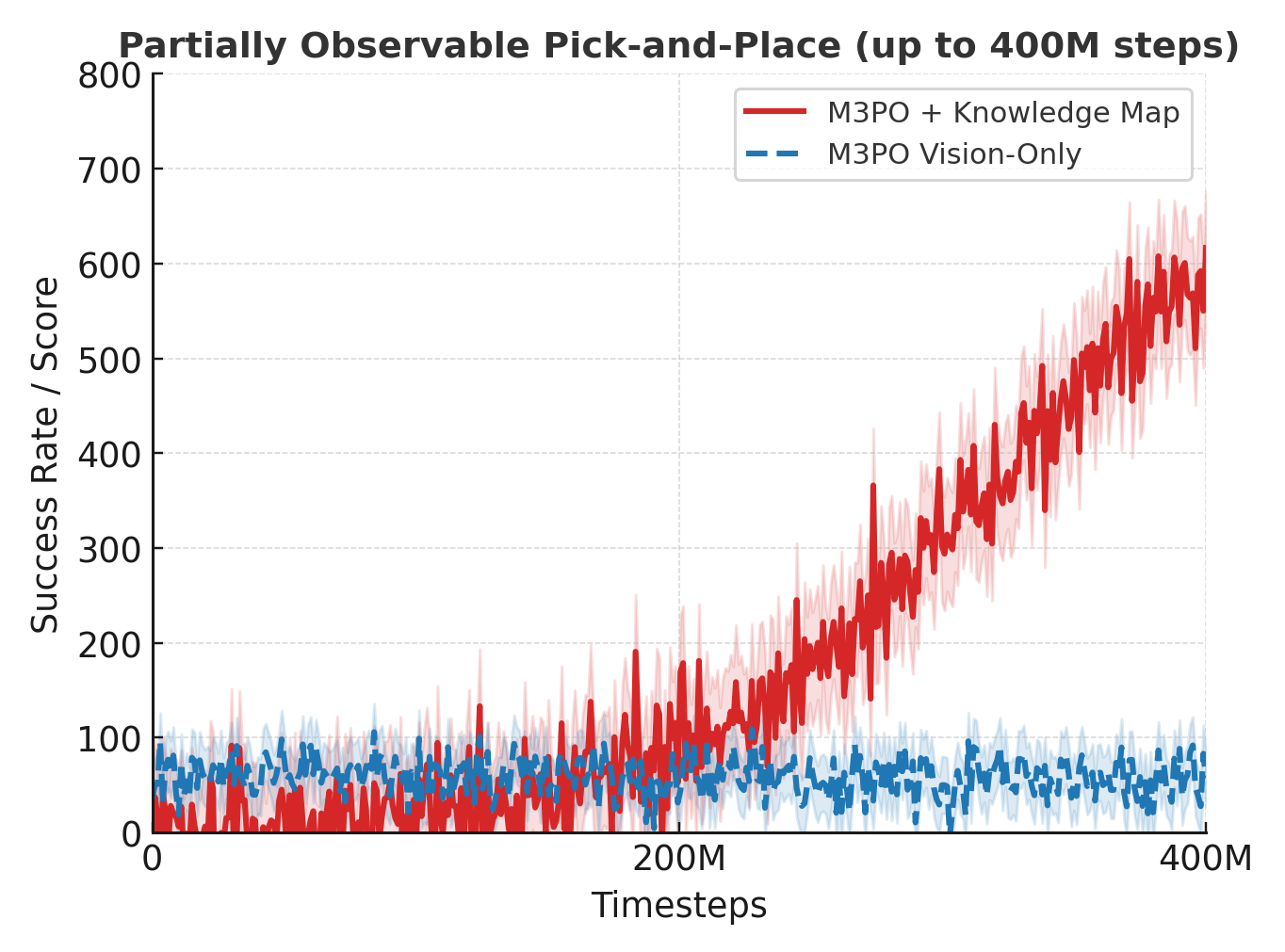}
    \caption{\textbf{PO Pick-Place.} M3PO (camera-only) vs.\ KG-M3PO (camera+KG).}
    \label{fig:po-m3po}
  \end{minipage}
\end{figure}

We train a \emph{single} policy on all tasks using \textbf{camera-only} observations (task ID concatenated), reporting the \emph{Normalized Multi-Task Score} (0–1000) over timesteps. As shown in Fig.~\ref{fig:mt-camera}, \textbf{M3PO} attains the highest score throughout training, followed by \textbf{IMPALA} and \textbf{TD-MPC2}; \textbf{MT-SAC} and \textbf{MT-PPO} learn more slowly. This ranking aligns with prior evidence that distributed actor–learner architectures can yield positive transfer in multi-task RL (IMPALA), while strong model-based agents scale well across diverse control tasks (for example TD-MPC2). 

\textbf{Protocol/selection.} We use this camera-only evaluation to pick the \emph{single best} multi-task baseline for subsequent partially observable experiments; accordingly, we carry forward \textbf{M3PO} and compare it against its knowledge-augmented counterpart \textbf{KG-M3PO} in order to isolate the benefit of the knowledge graph. 

\subsection{Partially Observable Tasks (M3PO vs.\ KG-M3PO)}
We compare the \textbf{M3PO} baseline (camera-only inputs) against \textbf{KG-M3PO} (camera inputs augmented with the online knowledge graph embedding) on two partially observable tasks referenced in Table~\ref{tab:tasks-small}. Fig.~\ref{fig:po-m3po} shows that adding the knowledge graph via KG-M3PO dramatically improves learning and final performance: KG-M3PO rises steadily, while the camera-only M3PO baseline remains near chance due to state aliasing under occlusion. This behavior is consistent with the role of dynamic scene graphs as a persistent relational state that retains task-relevant information beyond the current pixel observation.

\subsection{Qualitative Analysis: GPU Time vs.\ Knowledge Graph Utility}
\label{sec:qualitative-gpu}
\textbf{Compute vs.\ data efficiency.} We distinguish \emph{sample efficiency} (return vs.\ environment steps) from \emph{compute efficiency} (return vs.\ wall-clock/GPU-days). The KG stack adds per-step compute (detection, dynamic graph updates, graph encoding), so at a \emph{fixed} step budget it consumes more wall-clock than camera-only. However, KG-M3PO typically achieves target scores in \emph{fewer steps} (and often comparable wall-clock), especially in partially observable tasks where the camera-only M3PO baseline plateaus. Below we tabulate compute at a fixed step budget and a complementary \emph{time-to-target} view.

\begin{table}[t]
\centering
\begin{minipage}[t]{0.49\columnwidth}
\centering
\caption{Training cost (400M steps).}
\label{tab:cost-400m}
\scriptsize
\setlength{\tabcolsep}{2.5pt}
\renewcommand{\arraystretch}{1.05}
\begin{tabular}{l r r}
\toprule
Method & Time (s) & GPU-d \\
\midrule
M3PO (Cam)        & 52,734 & 0.610 \\
KG-M3PO (KG)  & 68,555 & 0.793 \\
\bottomrule
\end{tabular}
\end{minipage}
\hfill
\begin{minipage}[t]{0.49\columnwidth}
\centering
\caption{Efficiency to score 600.}
\label{tab:time-to-600}
\scriptsize
\setlength{\tabcolsep}{2.5pt}
\renewcommand{\arraystretch}{1.05}
\begin{tabular}{l r r}
\toprule
Method & Steps & GPU-d \\
\midrule
M3PO (Cam)        & N/A   & N/A \\
KG-M3PO (KG)  & 180M  & 0.36 \\
\bottomrule
\end{tabular}
\end{minipage}

\vspace{4pt}
\begin{minipage}{\columnwidth}
\begin{minipage}{\columnwidth}
\centering
{\footnotesize\textit{Cam} = camera-only; \textit{KG} = camera+knowledge graph.\\
N/A: target score not reached.}
\end{minipage}
\end{minipage}

\end{table}

Table~\ref{tab:cost-400m} reflects \emph{per-step} overhead (KG is slower per step). It does \emph{not} address how many steps each agent needs to reach a given score. KG-M3PO incurs higher \emph{per-step} cost (Table~\ref{tab:cost-400m}) but requires \emph{fewer steps} to attain competent performance (Table~\ref{tab:time-to-600}), yielding steeper return-vs-steps curves and robust gains on complex, partially observable tasks (occlusion, containment). For simple fully observable tasks under tight compute budgets, camera-only can be slightly faster early on; as complexity grows, KG’s structured state (visibility, containment, affordances) reduces perceptual aliasing and improves credit assignment, delivering higher asymptotic returns and better time-to-target despite overhead.

\subsection{Ablation: Robustness to Scene Graph Noise}
\label{subsec:ablation-noise}

Real-world deployment of the knowledge graph pipeline inevitably introduces
geometric uncertainty: detection and depth errors corrupt bounding-box
centers and extents, while imperfect pose estimation perturbs object
orientations.
To quantify how sensitive the policy is to such corruption, we conduct a
structured noise ablation on the \emph{Open-Cabinet} task (Franka) using
three independent noise axes applied to each graph node before it is passed
to the GNN encoder:

\begin{itemize}
  \item \textbf{Center noise.} Gaussian perturbation $\mathcal{N}(0,\sigma_c^2)$
        applied to each coordinate of the bounding-box center, where
        $\sigma_c = \alpha_c \cdot \bar{e}$ and $\bar{e}$ is the mean
        bounding-box extent; we vary $\alpha_c \in \{0\%, 2\%, 5\%\}$.
  \item \textbf{Extent noise.} Multiplicative Gaussian applied to each
        bounding-box dimension: $e' = e\,(1 + \epsilon)$,
        $\epsilon\!\sim\!\mathcal{N}(0,\sigma_e^2)$; we vary
        $\sigma_e \in \{0\%, 5\%, 10\%\}$.
  \item \textbf{Rotation noise.} Random SO(3) perturbation drawn from an
        isotropic Gaussian with standard deviation
        $\sigma_r \in \{0^\circ, 5^\circ\}$ applied via axis-angle
        composition.
\end{itemize}

\noindent
We evaluate three noise levels --- \textbf{Clean} ($\sigma=0$),
\textbf{Low} ($\sigma_c{=}2\%,\,\sigma_e{=}5\%,\,\sigma_r{=}5^\circ$), and
\textbf{High} ($\sigma_c{=}5\%,\,\sigma_e{=}10\%,\,\sigma_r{=}5^\circ$) ---
reporting final normalized score, success rate, and AUC.
All other hyperparameters and seeds are held fixed.

\begin{table}[h]
  \caption{\textbf{Noise ablation on Open-Cabinet (Franka, KG-M3PO).}
          Bounding-box noise (center, extent, rotation) degrades performance gracefully.}
  \label{tab:noise-ablation}
  \footnotesize
  \renewcommand{\arraystretch}{1.15}
  \setlength{\tabcolsep}{5pt}
  \begin{tabularx}{0.5\columnwidth}{@{} l c c c c @{}}
    \toprule
    \textbf{Noise level}
      & $\boldsymbol{\sigma_c}$
      & $\boldsymbol{\sigma_e}$
      & $\boldsymbol{\sigma_r}$
      & \textbf{Success (\%)} \\
    \midrule
    KG-M3PO -- Clean        & 0\%  & 0\%   & $0^\circ$    & 72 \\
    KG-M3PO -- Low noise    & 2\%  & 5\%   & $5^\circ$    & 68 \\
    KG-M3PO -- High noise   & 5\%  & 10\%  & $5^\circ$    & 61 \\
    \bottomrule
  \end{tabularx}

  \smallskip
  \noindent\footnotesize
  \textbf{Abbreviations:}
  $\sigma_c$ = center noise;
  $\sigma_e$ = extent noise;
  $\sigma_r$ = rotation noise.
\end{table}

The results in Table~\ref{tab:noise-ablation} show that the knowledge-conditioned
policy degrades gracefully: even under \emph{High} noise the KG-M3PO agent
(61\%) comfortably outperforms the camera-only M3PO baseline (50\%), confirming
that the GNN encoder learns representations robust to moderate geometric
uncertainty.
Center noise has the largest individual impact, as it directly shifts the
relational geometry used to compute spatial edges (\emph{in front of},
\emph{behind}, \emph{on}, \emph{under}); extent and rotation perturbations
produce a smaller but additive effect.
This robustness is partly attributed to the partial update strategy
(Sec.~\ref{subsec: knowledge-graph}): refreshing only the affected subgraph
every $n{=}10$ steps limits error accumulation from transient detection
failures.

\section{Conclusion and Future Work}

In this work, we demonstrated that \textbf{Perception $\rightarrow$ Knowledge $\rightarrow$ Policy works}: conditioning the policy on an online scene graph improves both sample efficiency and robustness under partial observability compared to the camera-only M3PO baseline. We also showed that \textbf{end-to-end matters}, as training the graph encoder (GNN) through the reinforcement learning loss aligns relational features with control and yields stronger long-horizon performance than detached or precomputed graph features. Furthermore, our results indicate that \textbf{multi-task scaling is feasible}, since a single policy can solve diverse manipulation tasks when the fused state representation exposes both low-level perception and high-level relational context.

Despite these contributions, several limitations remain. \textbf{Graph construction in the real world} is challenging due to detection errors, occlusions, and calibration drift, which may lead to performance degradation when graph quality decreases. \textbf{Highly dynamic settings} with rapid object motion, frequent contacts, and topology changes can disrupt temporal consistency of relations, reducing the usefulness of graph-derived context. Moreover, \textbf{no real-robot validation} has been performed: our results are limited to simulation, while on-hardware evaluation is complicated by latency, actuation constraints, and robust 3D scene-graph construction.

Looking forward, we plan to develop a \textbf{general graph interface} as a modular framework that supports different types of scene and knowledge graphs, including metric 3D graphs, affordance graphs, and open-vocabulary graphs, under a common API. Another important direction is \textbf{real-world deployment}, focusing on sim-to-real transfer and on-robot adaptation to evaluate the robustness of online graph construction and GNN updates on physical systems. We also aim to extend the approach to \textbf{highly dynamic environments}, equipping the pipeline with uncertainty-aware updates and recovery mechanisms for contact-rich manipulation and moving distractors. A further avenue is to explore \textbf{cross-embodiment generality} by training a single agent across different embodiments, such as Franka and UR5 with different grippers, through embodiment-conditioned fusion and shared graph semantics. Finally, we envision extending the framework to \textbf{mobile and loco-manipulation}, where navigation actions alter the scene graph and must be co-optimized jointly with manipulation.

\section*{ACKNOWLEDGMENT}
\thanks{*This work was supported by the Ministry of Economic Development of the Russian Federation (agreement No. 139-15-2025-013, dated June 20, 2025, subsidy identifier 000000C313925P4B0002)}

\bibliographystyle{IEEEtran}
\bibliography{reference}

\end{document}